\documentclass{IOS-Book-Article}

\usepackage{mathptmx}
\usepackage{soul}\setuldepth{article}
\usepackage[T1]{fontenc}
\def\hb{\hbox to 11.5 cm{}}

\usepackage{epsfig}
\usepackage{graphicx}
\graphicspath{{figures/}}
\usepackage{tikz}
\usepackage{pgfplots}
\usepackage{pgfplotstable}
\pgfplotsset{compat=1.16}
\usetikzlibrary{shapes,calc,decorations.markings,matrix,chains,positioning,decorations.pathreplacing,arrows,backgrounds,bending,quotes,arrows.meta,3d}
\definecolor{gray1}{rgb}{0.7, 0.7, 0.7}
\definecolor{gray2}{rgb}{0.8, 0.8, 0.8}
\definecolor{gray3}{rgb}{0.9, 0.9, 0.9}
\definecolor{red1}{rgb}{0.9, 0.4, 0.4}
\definecolor{red2}{rgb}{0.8, 0.3, 0.3}
\definecolor{red3}{rgb}{0.7, 0.2, 0.2}
\definecolor{green1}{rgb}{0.4, 0.7, 0.4}
\definecolor{green2}{rgb}{0.5, 0.8, 0.5}
\definecolor{green3}{rgb}{0.6, 0.9, 0.6}
\definecolor{blue1}{rgb}{0.5, 0.5, 0.7}
\definecolor{blue2}{rgb}{0.6, 0.6, 0.8}
\definecolor{blue3}{rgb}{0.7, 0.7, 0.9}
\definecolor{yellow1}{rgb}{1, 0.8, 0.6}
\definecolor{yellow2}{rgb}{1, 0.9, 0.6}
\definecolor{yellow3}{rgb}{1, 1, 0.6}

\definecolor{color0}{rgb}{1.0,0.1,0.1}
\definecolor{color1}{rgb}{1.0,0.3,0.3}
\definecolor{color2}{rgb}{1.0,0.5,0.5}
\definecolor{color3}{rgb}{1.0,0.7,0.7}
\definecolor{color4}{rgb}{0.1,0.1,1.0}
\definecolor{color5}{rgb}{0.3,0.3,1.0}
\definecolor{color6}{rgb}{0.5,0.5,1.0}
\definecolor{color7}{rgb}{0.7,0.7,1.0}

\usepackage{amsmath}
\usepackage{amssymb}
\usepackage{booktabs}
\usepackage{multirow}
\usepackage{caption} 
\usepackage{array}
\usepackage{xcolor, colortbl}

\usepackage{makecell}
\usepackage{arydshln}
\usepackage{subcaption}

\begin{document}

\pagestyle{headings}
\def\thepage{}
\begin{frontmatter}              

\title{Few-Shot Meta-Learning for Recognizing Facial Phenotypes of Genetic Disorders}

\markboth{}{} 

\author[A]{\fnms{Ömer} \snm{Sümer} \thanks{Corresponding Author: Ömer Sümer, oemer.suemer@informatik.uni-augsburg.de} },
\author[A]{\fnms{Fabio} \snm{Hellmann} },
\author[b]{\fnms{Alexander} \snm{Hustinx} },
\author[B]{\fnms{Tzung-Chien} \snm{Hsieh} },
\author[A]{\fnms{Elisabeth} \snm{André} },
and
\author[B]{\fnms{Peter} \snm{Krawitz}}

\runningauthor{Ö. Sümer et al.}
\address[A]{Chair for Human-Centered Artificial Intelligence, University of Augsburg}
\address[B]{Institute for Genomic Statistics and Bioinformatics, University of Bonn} 

\begin{abstract}
Computer vision-based methods have valuable use cases in precision medicine, and recognizing facial phenotypes of genetic disorders is one of them. Many genetic disorders are known to affect faces' visual appearance and geometry. Automated classification and similarity retrieval aid physicians in decision-making to diagnose possible genetic conditions as early as possible. Previous work has addressed the problem as a classification problem and used deep learning methods. The challenging issue in practice is the sparse label distribution and huge class imbalances across categories. Furthermore, most disorders have few labeled samples in training sets, making representation learning and generalization essential to acquiring a reliable feature descriptor. In this study, we used a facial recognition model trained on a large corpus of healthy individuals as a pre-task and transferred it to facial phenotype recognition. Furthermore, we created simple baselines of few-shot meta-learning methods to improve our base feature descriptor. Our quantitative results on GestaltMatcher Database show that our CNN baseline surpasses previous works, including GestaltMatcher, and few-shot meta-learning strategies improve retrieval performance in frequent and rare classes.
\end{abstract}

\end{frontmatter}

\markboth{}{} 

\section{Introduction}\label{section:introduction}
Genetic disorders affect more than 5\% of the population~\cite{Baird:1988}; in practice, physicians might fail to spot and clinically diagnose most of them. There is a set of genetic conditions and 30-40\% of them are known to affect craniofacial development and facial morphology~\cite{Hart:2009}. The alterations in the face and skull can be recognized by using computer vision. The output of computer vision-based systems can support physicians in diagnosing rare syndromes and eventually lead to therapeutic interventions.

The number of samples in real-life situations and databases shows considerable variation across disorders. This makes training deep convolutional networks not feasible, as in any object classification task. The nature of the problem necessitates addressing data imbalance and few-shot classification in facial phenotype analysis.

This paper presents an approach to improve the baseline for unseen facial genetic disorders based on a highly imbalanced distribution of disorders.

\section{Method}\label{section:method}
\begin{figure*}[ht!]
	\centering
	\def\svgwidth{0.8\linewidth}
	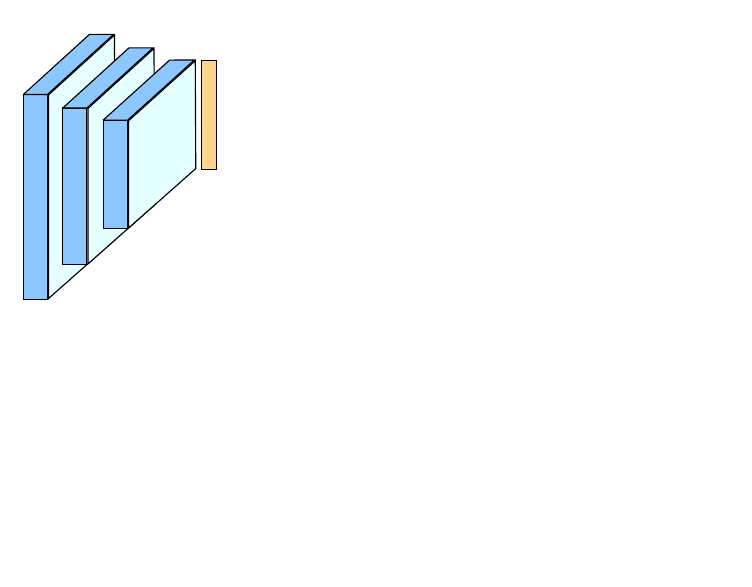
	\caption{Workflow of initial feature learning and few-shot meta-learning: top) the initial feature learning is done either on face recognition task or genetic disorder classification; bottom) the learned representation is used in few-shot meta-learning stage.}
	\label{figure:workflow}
\end{figure*}

Figure~\ref{figure:workflow} depicts the workflow of our proposed approach for facial phenotype recognition. The initial step in facial phenotype learning is to learn a solid initial representation. For this task, we trained a convolutional neural network backbone by adopting the metric learning-based Arcface loss~\cite{Deng:2019:Arcface} in face recognition.

After learning an initial representation, our approach is to learn a model from only a few annotated samples. Facial phenotypes for genetic diseases are highly imbalanced, and most categories have limited samples. Few-shot learning formulation can be seen as a meta-learning problem. There are separate support and query sets to learn to compare in the training and testing phases. These sets are created in an episodic manner, and K-way N-shot describes the task. The bottom part of Figure~\ref{figure:workflow} takes sampled episodes of support and query images and first extracts features using the backbone encoder by initializing from face recognition pre-trained weights. K-way means k different classes n images from each of them. During the training, the centroid of each embedding vector per class $c$ is calculated: $c_k = \frac{1}{\mathcal{S}_k} \sum_{x_i, y_i \in \mathcal{S}_k} f_\theta(x_i)$ where $x_i$ and $y_i$ are images and corresponding labels in each group of support set, $\mathcal{S}_k$.

Furthermore, differing from~\cite{Snell:2017:ProtoNets}, there are several K-way N-shot tasks in each episode. It refers to predicting the category of a query sample from K classes or N examples per class in the support set. This setting learns a feature embedding that can retrieve samples belonging to the same category using a similarity metric. The main difference here is that meta-learning is independent of the tasks and can better generalize on unseen classes.

In Prototypical Networks~\cite{Snell:2017:ProtoNets} the distance (or similarity) function is Euclidean distance. However, previous literature in facial phenotype recognition~\cite{Hsieh:2022:GestaltMatcher} used cosine similarity for the retrieval task. In order to make our meta-training as compatible as possible with our end task, we used cosine similarity between query embeddings, $f(x_i)$, and class centroids, $c_k$, and calculated logits as follows:
\begin{equation}
    p(y=k | x_i ) = \frac{e^{ \tau \cdot \langle f_\theta(x_i), c_k \rangle}}{\sum_k' e^{ \tau \cdot \langle f_\theta(x_i), c_k \rangle}}
\end{equation}
where $\tau$ is a learnable scalar that we applied to scale the values before applying the Softmax function following the related literature~\cite{Gidaris:2018,Chen:2021}.

We used version 1.0.3 of the GMDB~\cite{Hsieh:2022:GestaltMatcher} in our experiments. In version 1.0.3 the database contains 7.459 images of 449 syndromes in total. Genetic disorders are split into frequent and rare groups. In both datasets, faces are detected and aligned by RetinaFace~\cite{Deng:2019:Arcface}. Using five facial key points, we performed 5-point similarity alignment and normalized faces to the size of 112$\times$112. During the training of baseline classification and few-shot meta-learning models, we only applied channel mean and standard deviation normalization according to the train set statistics and random horizontal flipping.

In the training of both whole-set classification and few-shot meta-training, we used an SGD solver with a constant learning rate of 0.001 and weight decay of 0.0005 for 25 epochs. We used validation retrieval performance, specifically, the nearest neighbor retrieval of validation samples' feature embeddings to all training sets for model selection. The embedding size of the feature dimension is 512. During the few-shot meta-training, we sampled each episode containing four tasks, and the total number of episodes was kept at 100 and trained for 25 epochs.

We evaluated the performance of our classification and nearest-neighbor approach in terms of top-k accuracies in the frequent and rare test sets in the GMDB. Following Hsieh et al.~\cite{Hsieh:2022:GestaltMatcher}, learned facial embeddings were evaluated using three settings as follows: (1) the classification task reports only top-k accuracies using softmax outputs based on the frequent test set; (2) the retrieval task reports top-k accuracies using k-nearest neighbors based on feature embeddings and cosine distances from the frequent gallery and frequent test sets; and (3) the retrieval task reports top-k accuracies using k-nearest neighbors based on feature embeddings and cosine distances from the 10-Fold Cross-Validation rare gallery and rare test sets. We calculate the top-k accuracies for Top-[1, 5, 10, 30].

\section{Results}\label{section:results}
Table~\ref{table:ablation_results} depicts the results of our ablation study. As we aim to improve the retrieval performance on both tasks, we only evaluated GestaltMatcher DCNN using predictions trained with cross-entropy loss. The performance of GestaltMatcher DCNN trained on v1.0.3 of the database is aligned with the published results on \cite{Hsieh:2022:GestaltMatcher}. Top-1 accuracy varies in the ranges of 15\% to 21\% in frequent and rare sets where the total number of classes is 204 and 245, respectively. Our stronger baseline, a ResNet-50 trained on MS1MV2 using ArcFace loss (Enc-healthy), performed 34.06\% top-1 accuracy in the frequent set, whereas GestalthMatcher DCNN's retrieval performance remains at 15.96\%.

\begin{figure}
    \centering
    \begin{minipage}[b]{0.49\textwidth}
        \centering
        \captionsetup{type=table}
        \resizebox{\textwidth}{!}{
        \begin{tabular}{llllll}
            \toprule
             Method  & Top-1 & Top-5 & Top-10 & Top-30 \\ 
            \midrule
            \textbf{Frequent set}  &            &       &       &        &        \\  
            \midrule
            \multicolumn{2}{l}{GestaltMatcher DCNN~\cite{Hsieh:2022:GestaltMatcher}}  &        &       &  \\  
            Classification  & 21.21 & 42.08 & 54.60 & 73.92 \\
            Retrieval       & 15.96 & 33.83 & 45.46 & 69.64 \\ \hdashline
            Enc-healthy     & 34.06 & 53.96 & 64.42 & \textbf{81.28} \\
            Enc-all  (GMDB)  & \textbf{42.50} & 58.18 & 65.26 & 78.08 \\
            Enc-base (GMDB)  & 40.47 & \textbf{60.71} & \textbf{67.29} & 79.09 \\
            \midrule
            \textbf{Rare set}  &            &       &       &        &        \\  
            \midrule
            \multicolumn{2}{c}{GestaltMatcher DCNN~\cite{Hsieh:2022:GestaltMatcher}}  & & &   \\  
            Retrieval     & 19.26 & 36.28 & 44.07 & 60.73 \\ \hdashline
            Enc-healthy   & 26.31 & 42.62 & 46.98 & 62.92 \\
            Enc-all (GMDB) & 26.40  & 42.36 &	50.42 &	65.76 \\
            Enc-base (GMDB) & \textbf{28.25} & \textbf{44.88} &	\textbf{52.00} & \textbf{66.18} \\
            \bottomrule
        \end{tabular}
        }
        \caption{Performance comparison of GestaltMatcher DCNN and our baseline models on GMDB (v1.0.3).}
        \label{table:ablation_results}
    \end{minipage}
    \hfill
    \begin{minipage}[b]{0.49\textwidth}
        \centering
        \captionsetup{type=table}
        \resizebox{\textwidth}{!}{
        \begin{tabular}{lllll}
            \toprule
             Method  & Top-1 & Top-5 & Top-10 & Top-30 \\ 
            \midrule
            \textbf{Frequent set}  & &  & & \\  
            \midrule
            GMDB-fs          & 48.06 & 68.13 & 75.89 & 85.67 \\  
            \multicolumn{2}{l}{\emph{(feature-level fusion)}}  & & &   \\ \hdashline
            +Enc-healthy     & 47.55 & \textbf{68.47} & \textbf{77.23} & \textbf{88.69} \\  
            +Enc-all (GMDB)  & \textbf{47.55} & 67.62 & 74.20 & 84.65 \\  
            +Enc-base (GMDB) & 47.22  & 67.96  & 74.71 & 84.82 \\ 
            \midrule
            \textbf{Rare set}    &    &   &   &        \\  
            \midrule
            GMDB-fs  & 30.21 &	48.19 &	56.39 &	71.07   \\
            \multicolumn{2}{l}{\emph{(feature-level fusion)}}  & & &   \\ \hdashline 
            +Enc-healthy     & 32.89 & \textbf{50.65} & \textbf{57.89} & \textbf{71.39} \\  
            +Enc-all (GMDB)  & 30.88 & 48.29 & 56.57 & 70.54 \\  
            +Enc-base (GMDB) & \textbf{33.08} & 48.37 & 56.65 & 70.72 \\ 
            \bottomrule
        \end{tabular}
        }
        \caption{Few-shot meta baseline and feature-level fusion on GMDB (v1.0.3) retrieval task.}
        \label{table:few_shot_meta}
    \end{minipage}
\end{figure}

Few-shot meta baseline that we adopted in our experiments is a 10-way 3-shot task with 2 query samples in each task. Following~\cite{Chen:2021}, we sampled multiple tasks in each episode. The reported experiments are done using four tasks per episode. Table~\ref{table:few_shot_meta} shows the retrieval performance of few-shot meta-learning models on both frequent and rare test sets.

Few-shot meta-training (GMDB-fs) improves the top-1 frequent test accuracy of the best GMDB-trained baseline models, Enc-all and Enc-base by 7.59\%, and 5.56\%, respectively. This improvement is not limited to top-1 retrieval, it is also retained in different neighbor retrieval. We initialized GMDB-fs models using healthy encoding.

\begin{table*}[ht!]
    \centering
    \begin{tabular}{lllllllll}
        \toprule
         & \multicolumn{4}{c}{Frequent} & \multicolumn{4}{c}{Rare} \\
         & Top-1 & Top-5 & Top-10 & Top-30 & Top-1 & Top-5 & Top-10 & Top-30 \\ 
        \midrule
        n-categories & \multicolumn{8}{l}{3-shot / 2-query} \\
        \midrule
        5   & \textbf{49.24} & 66.44 & 75.04 & 84.49 & 27.70 & 45.41 & 54.44 & 69.37 \\
        10  & 48.06 & \textbf{68.13} & \textbf{75.89} & 85.67 & 30.21 & 48.19 & 56.39 & 71.07 \\
        15  & 47.89 & 67.96 & 75.72 & \textbf{86.51} & \textbf{31.63} & \textbf{49.35} & \textbf{58.17} & \textbf{72.95}\\
        20  & 48.06 & 67.62 & 74.37 & 84.65 & 27.76 & 47.04 & 55.29 & 69.33 \\
        \midrule
        n-shot / n-query & \multicolumn{8}{l}{10-categories} \\
        \midrule
        1 / 4  & 44.35 & 65.94 & 73.19 & 84.65 & 29.37 & 46.99 & 56.26 & 69.47\\
        2 / 3  & 44.35 & 67.12 & 74.87 & \textbf{86.34} & \textbf{31.77} & \textbf{49.26} & \textbf{57.66} & 71.04 \\
        3 / 2  & 47.05 & \textbf{69.14} & \textbf{76.05} & \textbf{86.34} & 30.15 & 48.38 & 57.17 & \textbf{71.34}\\
        4 / 1  & \textbf{48.23} & 68.13 & 75.21 & 84.65 & 27.76 & 45.79 & 55.58 & 68.72\\
        \midrule
    \end{tabular}
    \caption{Comparison of n categories with 4 tasks per episode and 10 categories with n-shot and n-query.}
    \label{table:configuration_comparison}
\end{table*}

In both frequent and rare sets, feature-level fusion with the healthy encoder performed the best in nearly all retrieval tasks. In top-1 rare retrieval, fusion with Enc-base gives the best accuracy, 33.08\%. Even though Enc-all and Enc-base perform better than Enc-healthy, their performance on feature fusion is limited. 

\section{Discussion}\label{section:discussion}
We observed differences in the model's behavior when evaluating the few-shot meta-based training with different sets of configurations (Table~\ref{table:configuration_comparison}). These variables affect the difficulty of few-shot tasks and need to be examined in depth. One of them is the number of ways to define possible classes in a support set. When the number of categories increases in episodic training, in ranges of  [5, 10, 15, 20], the best retrieval performance is acquired with 10-way and 15-way. We consider this behavior related to the complexity of classification tasks in each episode.

Top-k accuracies in frequent and rare sets are better at 10-way and 15-way. We picked the 10-way setting to evaluate other parameters that affect the performance of episodic training. These are the number of images in each class in the support set (k-shot) and the number of query images. The bottom part of Table~\ref{table:configuration_comparison} presents evaluation performance using the different number of shots and queries. 
A higher number of shots improves frequent set retrieval performance; however, the 4-shot setting performs worse in rare set retrieval. We argue that the minimum number of shots must be descriptive enough according to the n-way task learned. In our experiments, we could not increase n-shot and queries more as the minimum number of samples per class in the training set was 5. It can be seen in Table~\ref{table:configuration_comparison} that 2-shot or 3-shot settings give the best overall performance.

\section{Conclusion}\label{section:conclusion}
In this work, we addressed facial phenotype recognition for genetic disorders. We tackled the issue of data scarcity and imbalanced data with a few-shot meta-based learning approach. Therefore, we trained a state-of-the-art face recognition model on standard face recognition databases. The learned facial representations were then transferred to our low-resource target data domain, the GestaltMatcher Database. With the few-shot meta-learning, we improved genetic disorder recognition on mainly unseen disorders compared to the recently published GestaltMatcher DCNN.
In future work, using generative models on either image or feature level, synthesized samples can be added to few-shot training and reduce the effect of uneven class distribution.

\bibliographystyle{vancouver.bst}
\bibliography{MIE2023_references}

\end{document}